\documentclass[conference]{IEEEtran}
\usepackage{times}

\IEEEoverridecommandlockouts
\usepackage{graphicx}
\def\BibTeX{{\rm B\kern-.05em{\sc i\kern-.025em b}\kern-.08em
    T\kern-.1667em\lower.7ex\hbox{E}\kern-.125emX}}
\usepackage[font=small,labelfont=bf,tableposition=top]{caption}

\usepackage[numbers]{natbib}
\usepackage{multicol}
\usepackage[bookmarks=true]{hyperref}
\usepackage{hyperref}

\usepackage{capt-of,etoolbox}

\hypersetup{
    colorlinks=true,
    linkcolor=blue,
    filecolor=magenta,      
    urlcolor=blue,
    pdftitle={Overleaf Example},
    pdfpagemode=FullScreen,
    }

\usepackage{graphics} 
\usepackage{amsmath} 
\usepackage{amssymb}  
\usepackage{colortbl}
\usepackage{color}
\usepackage{mathtools}
\usepackage{subfigure}
\usepackage{amsmath,amssymb,amsfonts}
\usepackage{algorithm}
\usepackage{amssymb}  
\usepackage{algorithmic}
\usepackage{graphicx}
\usepackage[dvipsnames]{xcolor}
\usepackage{tabularx}
\usepackage{multirow}
\usepackage{colortbl}
\usepackage{color}

\usepackage[belowskip=-15pt,aboveskip=0pt]{caption}

\newcommand{\loosepar}{\looseness=-1}

\begin{document}

\title{LIVE: Lidar Informed Visual Search for Multiple Objects with Multiple Robots}


\author{\authorblockN{Ryan Gupta\authorrefmark{1},
Minkyu Kim\authorrefmark{2},
Juliana T Rodriguez\authorrefmark{2}, 
Kyle Morgenstein\authorrefmark{1} and
Luis Sentis\authorrefmark{1}}
\thanks{\authorrefmark{1}Department of Aerospace Engineering and Engineering Mechanics,
        University of Texas at Austin,
        Austin, TX 78712 USA
        {\tt\small ryan.gupta@utexas.edu}}%
\thanks{\authorrefmark{2}Department of Mechanical Engineering,
        University of Texas at Austin,
        Austin, TX 78712 USA
}}

\let\oldtwocolumn\twocolumn
\renewcommand\twocolumn[1][]{%
    \oldtwocolumn[{#1}{
    \begin{center}
    \vspace{-3em}
           \includegraphics[width=\textwidth]{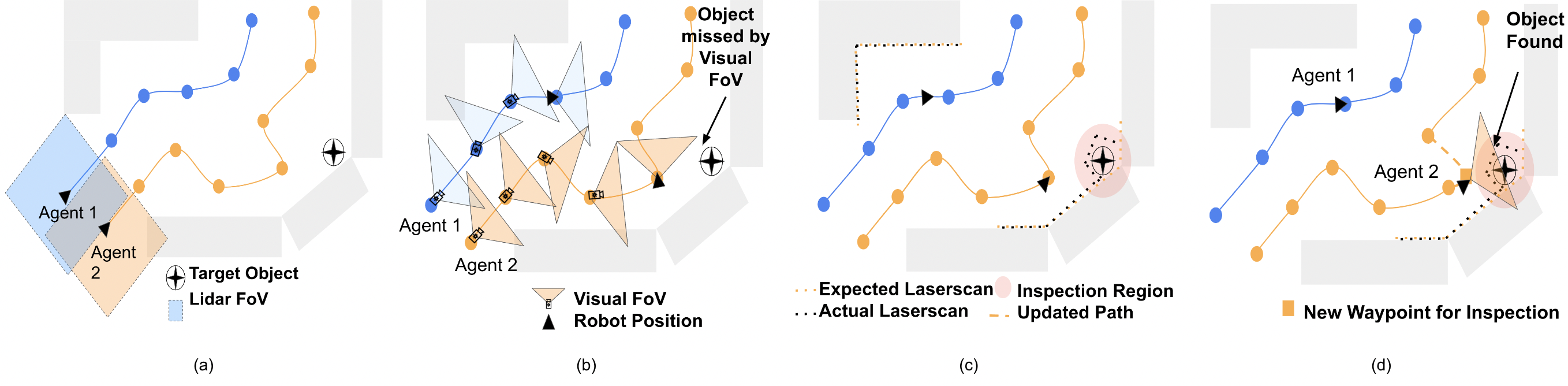}
           \captionsetup{belowskip=1pt}
           \captionof{figure}{An overview of LIVE. In (a), global path plans are generated for Lidar CPP. (b) While following the global paths without LIVE, the agent may miss the object of interest due to a differently sized camera and lidar FoV. (c) During localization, if an unmapped object, represented by a black circle with a cross, is discovered using lidar segmentation, it is labeled as a inspection region. (d) The Waypoint Manager (Sec. \ref{sec:wpt-mgr}) generates an intermediate waypoint for visual inspection of the unexpected object.\loosepar{}}
           \label{fig:curiosity}
        \end{center}
    }]
}
\smallskip

%

\maketitle
\IEEEpeerreviewmaketitle

\begin{abstract}

This paper introduces LIVE: Lidar Informed Visual Search focused on the problem of multi-robot (MR) planning and execution for robust visual detection of multiple objects. 
We perform extensive real-world experiments with a two-robot team in an indoor apartment setting. 
LIVE acts as a perception module that detects unmapped obstacles, or Short Term Features (STFs), in Lidar observations.
STFs are filtered, resulting in regions to be visually inspected by modifying plans online. 
Lidar Coverage Path Planning (CPP) is employed for generating highly efficient global plans for heterogeneous robot teams.
Finally, we present a data model and a demonstration dataset, which can be found by visiting our project website https://sites.google.com/view/live-iros2023/home.\loosepar{}

\end{abstract}

\section{Introduction}
Autonomous planning and real-world execution for multi-robot (MR) CPP, Active Object Search, and Exploration are receiving significant attention from the robotics community due to their relevance in several real-world scenarios including cleaning, lawn mowing, inspection, surveillance, and SAR \cite{tan2021comprehensive}.
This paper addresses efficient and robust path planning for real-world MR teams performing multi-object visual detection by combining Coverage Path Planning (CPP) with active sensing. 
 The goal in CPP is to generate path plans such that a sensor footprint, or Field of View (FoV), covers the region of interest \cite{choset2005principles}. 
Efficiency is commonly measured by coverage time, path length or FoV overlap \cite{tan2021comprehensive}. 
Early work \cite{ahmadi2005continuous} consider various cost functions for continuous and varying-rate area sweeping with a single robot and multiple robots \cite{ahmadi2006multi}.
MR CPP offers several benefits including speed and resilience to robot failure and is a common feature in search and rescue (SAR) and other critical applications.
State of the art work in multi-aerial vehicle exploration in confined spaces \cite{best2022resilient} consider the problem of multi-sensor exploration for the purpose of visual mapping or inspection of surfaces.
They leverage range scan data for guiding visual surface exploration. Instead, we propose to leverage range scan information for guiding visual search.
Active Sensing was first proposed in \cite{bajcsy1988active} as a method of providing a control strategy based on current world state, updated by observations, and is frequently employed in information gathering missions.
Active sensing has proven to be an effective tool for information gathering missions including SLAM \cite{charrow2015information}, search \cite{aydemir2013active,ye1996sensor}, and object tracking \cite{kim2022active}.
In the MR setting, \citet{gosrich2021coverage} enable robot teams position themselves to observe events using a Graph Neural Network for non-local information during decision making. 
In \cite{igoe2021multi}, MR multi-object search in unknown environments is cast as a reinforcement learning problem and address non-myopic planning.\loosepar{}



While MR CPP offers several benefits, computing optimal paths for multiple agents is NP-Hard \cite{guo2022sub}.
It remains an ongoing problem to improve path efficiency and computational load in MR CPP and exploration \cite{tan2021comprehensive}.
\citet{kim2022coverage} provide high efficiency plans by combining sampling and optimization.
However, in visual CPP with a small FoV RGB camera, it becomes impractical to plan online due to increased visitation sites.
Furthermore, \cite{drew2021multi} notes the lack of multi-robot systems deployed for real-world autonomous search, even with increased publications.
They indicate a need for research focused on the realistic evaluation of methods for real-world search. \loosepar{}

Mobile robots are frequently equipped with Lidar, which cast a significantly wider FoV than an RGB camera. 
LIVE leverages the wide FoV Lidar sensor to inform visual inspection in real-world experiments.
LIVE classifies incoming Lidar observations to remove dynamic features and static map obstacles leaving Short Term Features (STFs), defined as static, unmapped obstacles.
Raw STFs are filtered into inspection regions, defined as possible target object locations.
Agents select among inspection regions, inspect them visually, then continue along global plans.
An overview is shown in Fig. \ref{fig:curiosity}.
This online modification of global plans enables the fast planning of efficient paths based on wide FoV Lidar scans at the global stage, while still managing robust visual results from the active sensing approach. \loosepar{}


The contributions of this work can be summarized as follows:\loosepar{}
\begin{itemize}
    \item Propose a new method for incorporating lidar information to real-world multi-robot multi-object visual search  \loosepar{}
    \item Uniquely combine Lidar CPP and Active Sensing for visual object search
    \item Deploy extensively in a heterogeneous two-robot system for verification and baseline comparison \loosepar{} 
    
\end{itemize}

\section{Methods}
\label{sec:methods}

This paper focuses on efficient and robust multi-robot multi-object detection in indoor environments with static map information known. 
The approach leverages global multi-robot CPP for efficient global plans and a perception module capable of modifying those plans online.
An overview can be seen in Fig. \ref{fig:curiosity}. First, global path plans are generated for heterogeneous multi-robot teams \cite{kim2022coverage}.\loosepar{}


\subsection{Search Map (Entropy Map)}
\label{sec:search-map}
 Bayesian filtering is employed to maintain a target estimate over the map. 
 Each cell is assigned a probability of occupancy between $0$ and $1$. 
 A cell is initialized to $0.5$, except those corresponding to the static map, which are assigned $1$. 
 Cells that have been visited by the sensor FoV and are free are assigned $0$. 
 The search map is updated at each step when the central server receives local costmap observations from each robot that represent sensor FoV.
 In this work, local costmaps are rectangular, representing Lidar FoV, or triangular, representing visual FoV.
 Entropy over the map is measured by considering all cells in the 2D global costmap and is computed as
\begin{equation}
 H(M_t) = -\sum_{i=1}^{N}(m^i_t\log(m^i_t)+(1-m_t^i)log(1-m_t^i))
 \end{equation}
 where $m^i_t$ is the occupancy variable at time step $t$ and $N$ denotes the total number of cells. For further details refer to \cite{kim2022coverage}. \loosepar{}


\subsection{Inspection Region Detection}
\label{sec:curiosity-detection}
Computing inspection regions begins with the assumption that the objects of interest belong to the set of unmapped obstacles.
Incoming Lidar observations are classified based on current robot pose estimate.
Each point in the 2D scan is classified as a Long Term Feature (LTF), Short Term Feature (STF), or Dynamic Feature (DF) \cite{biswas2017episodic}.
LTFs represent the static map obstacles and STFs represent static unmapped obstacles.
Let $x_i$ denote the pose of the robot, and $s_i$ denote observation at time step $t_i$. Each observation $s_i$ consists of $n_i$ 2D points, $s_i=\{p^j_i\}_{j=1:n_i}$. 
Observations are transformed from robot local frame into the global frame using an affine transformation $T_i\in SE(3)$. 
Finally, let map $M$ be represented as a set of lines $\{l_i\}_{1:n}$.\loosepar{}

\subsubsection{LTF}

First, an analytic ray cast is performed \cite{biswas2012depth} to determine expected laserscan based on map $M$ and current robot position $x_i$.
Given observations, the probability that points correspond to one of the lines of that static map can be written \loosepar{}
 \begin{equation}
 \label{eq:ltf}
 P(p_i^j|x_i,M) = exp\left(-\frac{dist(T_ip_i^j,l_j)^2}{\Sigma_s}\right)
  \end{equation}
where $\Sigma_s$ is the scalar variance of observations, which comes from sensor accuracy. If Eq. \ref{eq:ltf} is greater than a threshold, point $p_i^j$ is classified as a LTF.
  
\subsubsection{STF}
Remaining points will be classified as STF or DF. 
Observations at current time $i$, $p_i^j$, are compared with prior observations at time $k$, $p_k^l$ to determine correspondence between points in subsequent observations. 
The likelihood of the remaining points corresponding to the same point as in a previous laserscan is computed as  
 \begin{equation}
\label{stf_eqn}
 P(p_i^j,p^l_k|x_i,x_k) = exp\left(-\frac{||T_ip_i^j-T_kp^l_k||^2}{\Sigma_s}\right)
  \end{equation}
 where $p^k_l$ is the nearest point from $p_i^j$ among points which does not belong to LTF at other timesteps, defined as
 \begin{equation}
 p^l_k = \arg min ||T_ip_i^j-T_kp^l_k||
 \end{equation}
When Eq. \ref{stf_eqn} is greater than some threshold, point $p_i^j$ is classified as an STF.
Remaining points in $s_i$ are classified as DFs, which are ignored in this study.

\begin{figure}[t]
    \centering
    \includegraphics[width=\columnwidth]{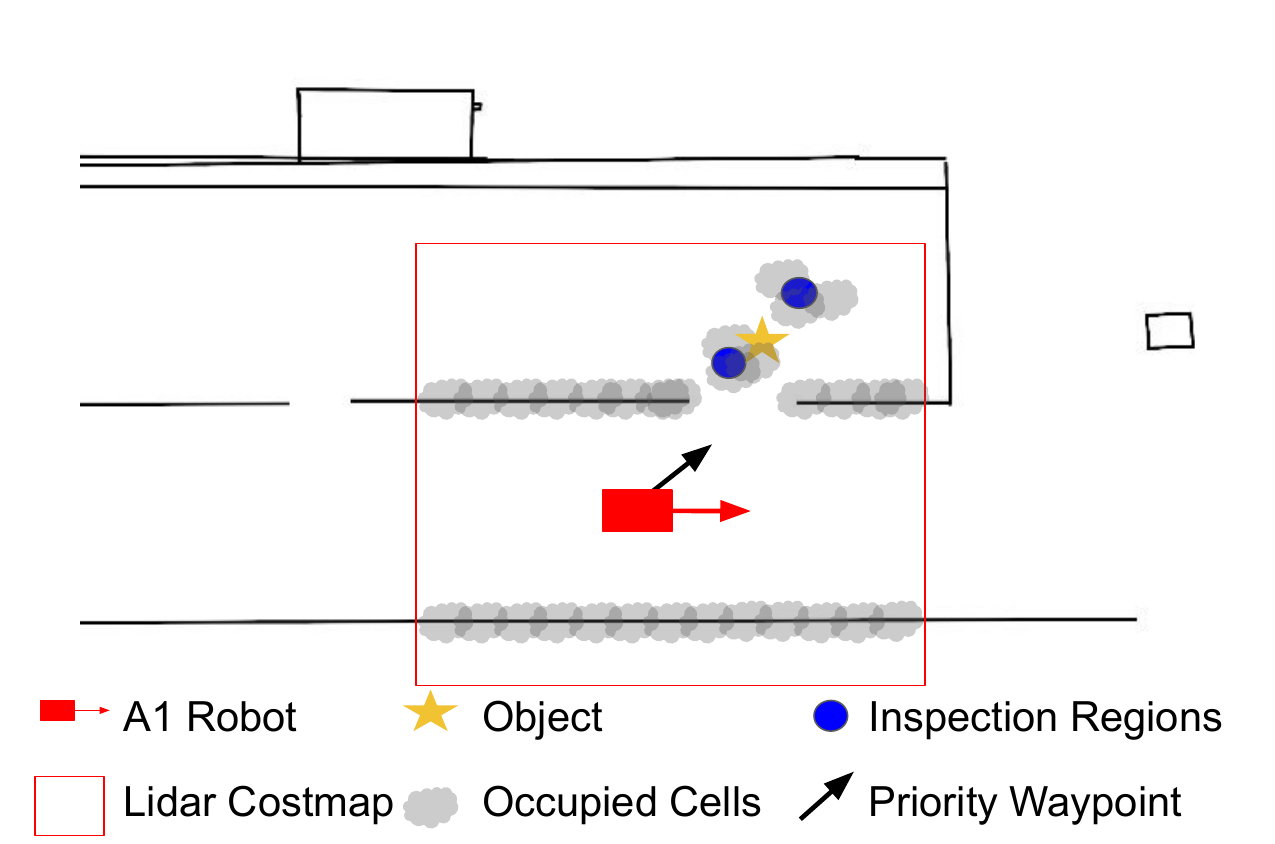}
    \vspace{0.25em}
    \caption{The lidar costmap is shown over a portion of the map to demonstrate the Lidar FoV detecting unmapped obstacles. Unmapped obstacles, or STFs, are filtered into inspection regions. This figure occurs the moment A1 is given a priority waypoint from LIVE for viewing an inspection region.\loosepar{}
    \label{fig:trial5-a1}}
\end{figure}

\subsubsection{Inspection Region Selection}
STFs obtained in the previous subsection are generated stochastically using the entire set of Lidar points from estimated robot pose.
As a result, the set of raw STFs is large and filtering steps are critical.
First, the pooling operator is used to reduce duplicates within a radius.
Second, pooled STFs within a certain distance of LTFs are removed to eliminate false positives caused by localization drift.
Finally, points inside visually observed regions of the map are removed.
The remaining points are inspection regions.
During waypoint generation the nearest inspection region is selected for generating a priority waypoint.
The result of this process is shown in Fig. \ref{fig:trial5-a1}. \loosepar{}

\subsection{Waypoint Manager}
\label{sec:wpt-mgr}
The Waypoint Manager takes global paths as input and acts as a finite state machine to determine when to send the robot an updated navigation waypoint. 
This node also receives priority waypoints for viewing inspection regions at each localization timestep.
Due to the high update frequency and the stochastic nature of inspection regions, exhaustively visiting all priority waypoints is inefficient.
As a result, this node incorporates priority waypoints every so often if they are available.
The maximum rate of occurrence of priority waypoints is a tunable parameter that will impact robustness and target detection time.
After visiting the priority waypoint, the manager will resume along the global path plans.
A priority waypoint being selected is shown in Fig. \ref{fig:trial5-a1}.

\begin{figure}[t]
    \centering
    \includegraphics[width=0.9\columnwidth]{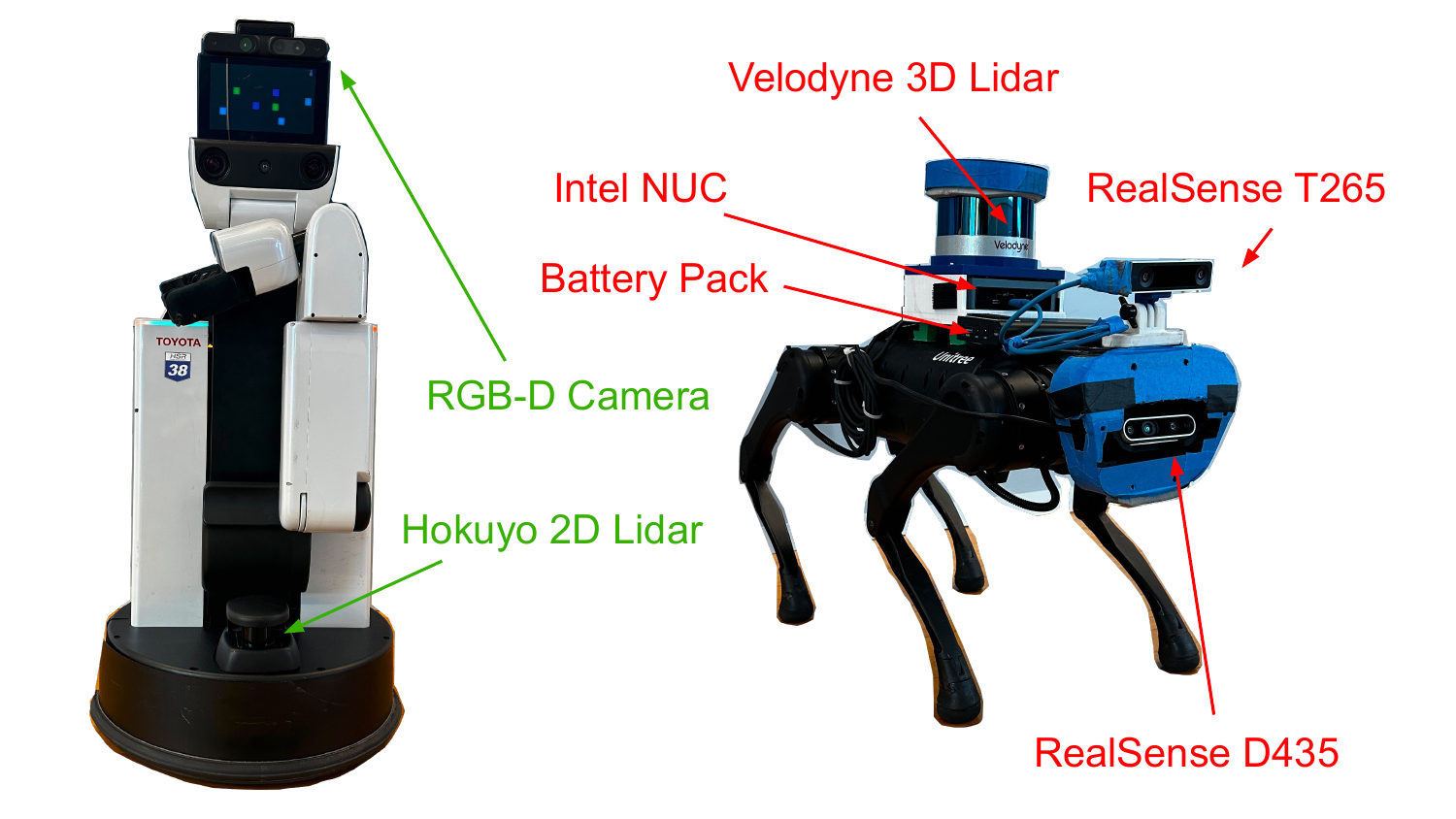}
    \caption{Figure of two robot team with key components labeled.}
    \label{fig:robots}
\end{figure}

\section{Task Description and Experiment Setup}
\label{sec:setup}

\begin{figure}[t!]
    \centering
    \includegraphics[width=\columnwidth]{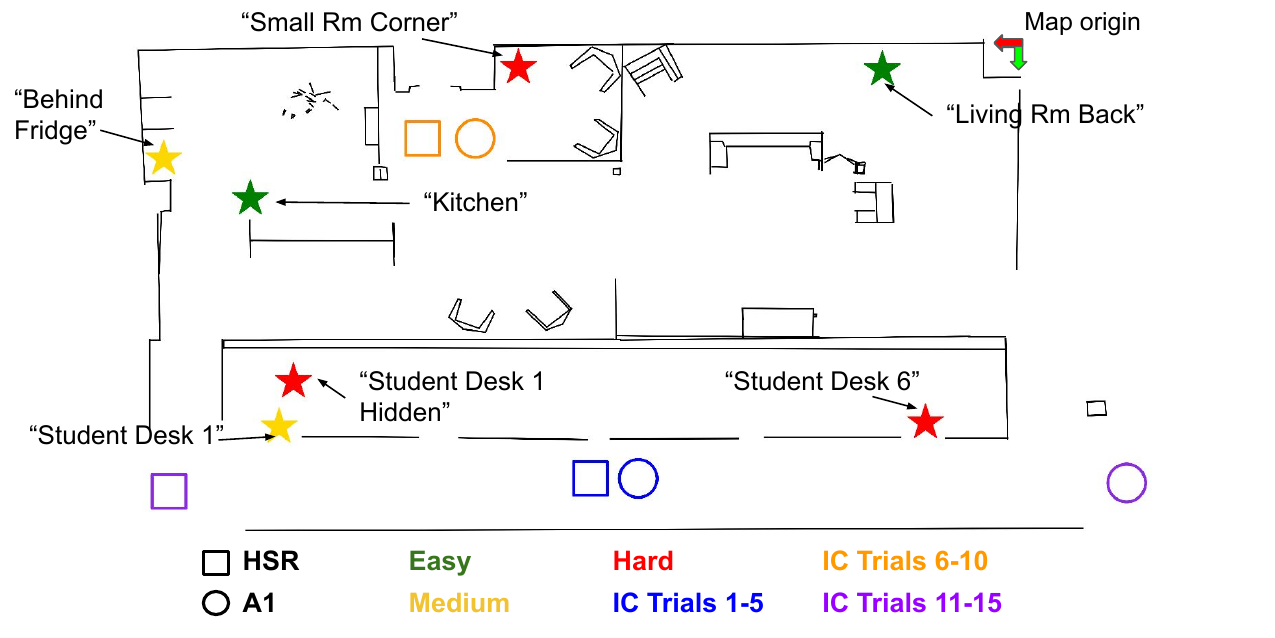}
    \caption{Top view of the static map with each of the three robot initial conditions (IC) and seven object location labeled. A1 ICs are circles and HSR ICs are squares, color-coordinated for the trials. The object locations are shown by stars whose color indicates difficulty with location names included.\loosepar{} }
    \label{fig:map-ic-targets}
    \vspace{0.5em}
\end{figure}

A team comprised of the Unitree A1 Quadruped \cite{unitree-A1} and the Toyota HSR \cite{yamamoto2019development} must visually detect two static objects of interest. 
Specifically, robots must find two small suitcases in a 20mx30m apartment setting with an attached hall. 
The goal of the team is to robustly detect objects where efficiency is measured as path length.
Robots with their sensors and instrumentation are described in Fig. \ref{fig:robots}.
The HSR employs Toyota move base to generate movement commands from given waypoints. 
The A1 leverages a carrot planner to navigate to waypoints.
LIVE and waypoint manager nodes run aboard each robot.
Two maps are maintained: the first is the Search Map, which is implemented as a 2D global costmap in ROS and the second is a vectormap used for localization \cite{biswas2017episodic}, shown in Fig \ref{fig:map-ic-targets}.
Both maps can be found alongside the dataset at the project website.
A central search server generates path plans and maintains the search map  
by sending global plans to robots and receiving position and costmap information back.
Communication between the central server and each robot are performed using Robofleet \cite{sikand2021robofleet}.
A local network covers the full region using a ASUS AC1900 WiFi router. 
The laptop and onboard computers run Ubuntu 18.04 and ROS Melodic \cite{ros} to implement all of the capabilities. 
Three planner settings are compared: $1)$ Lidar CPP, $2)$ Heuristic Visual CPP, and $3)$ Lidar CPP + LIVE
The three planner settings are described in Section \ref{sec:methods}.
In each of the three planning settings, 15 trials performed, for a total of 30 potential objects to be found.
The 15 trials are composed of five trials from each of three different initial conditions (IC), depicted on the static map in Fig. \ref{fig:map-ic-targets}.
Object locations change between trials with varying difficulty. The same five sets of object locations are tested from each IC.
There are seven total object locations used, categorized as easy, medium, or hard and they are also shown in Fig. \ref{fig:map-ic-targets}.\loosepar{}

\section{Results \& Discussion}
\label{sec:discussion}

\begin{table}[]
\centering
\begin{tabular}{|c||c|c|c|} 
\hline
\multicolumn{4}{|c|}{\textbf{Heuristic Visual CPP}} \\
\hline
  & \textbf{HSR Avg.}  & \textbf{A1 Avg.} & \textbf{Overall Avg.}  \\
\hline
\textbf{IC1} & $ 36.9 $  & $ 46.5 $ & $ 41.7 $ \\ 
\hline
\textbf{IC2} & $ 15.0 $  & $ 32.2 $ & $ 23.6 $ \\ 
\hline
\textbf{IC3} & $ 30.2 $  & $ 44.1 $ & $ 37.2 $ \\ 
\hline
 & $ 27.4 $ & $ 40.9 $ & $ \color{BrickRed} \textbf{34.2} $  \\ 
\hline
\hline
\multicolumn{4}{|c|}{\textbf{Lidar CPP + LIVE (Our Method)}} \\
\hline
  & \textbf{HSR Avg.}  & \textbf{A1 Avg.} & \textbf{Overall Avg.}  \\
\hline
\textbf{IC1} & $ 19.4 $  & $ 32.9 $ & $ 26.1 $ \\ 
\hline
\textbf{IC2} & $ 17.0 $  & $ 22.5 $ & $ 19.7 $ \\ 
\hline
\textbf{IC3} & $ 25.2 $  & $ 46.0 $ & $ 35.6 $ \\ 
\hline
 & $ 20.5 $ & $ 33.8 $ & \color{BrickRed} $ \textbf{27.2} $  \\ 
\hline
\end{tabular}
\captionsetup{belowskip=0pt}
\caption{Table showing average path lengths for each robot and combined in meters as a function of initial condition.}
\vspace{-2.5em}
\label{tab:lengths}
\end{table}
\begin{figure*}[]
    \centering
    \includegraphics[width=1.5\columnwidth]{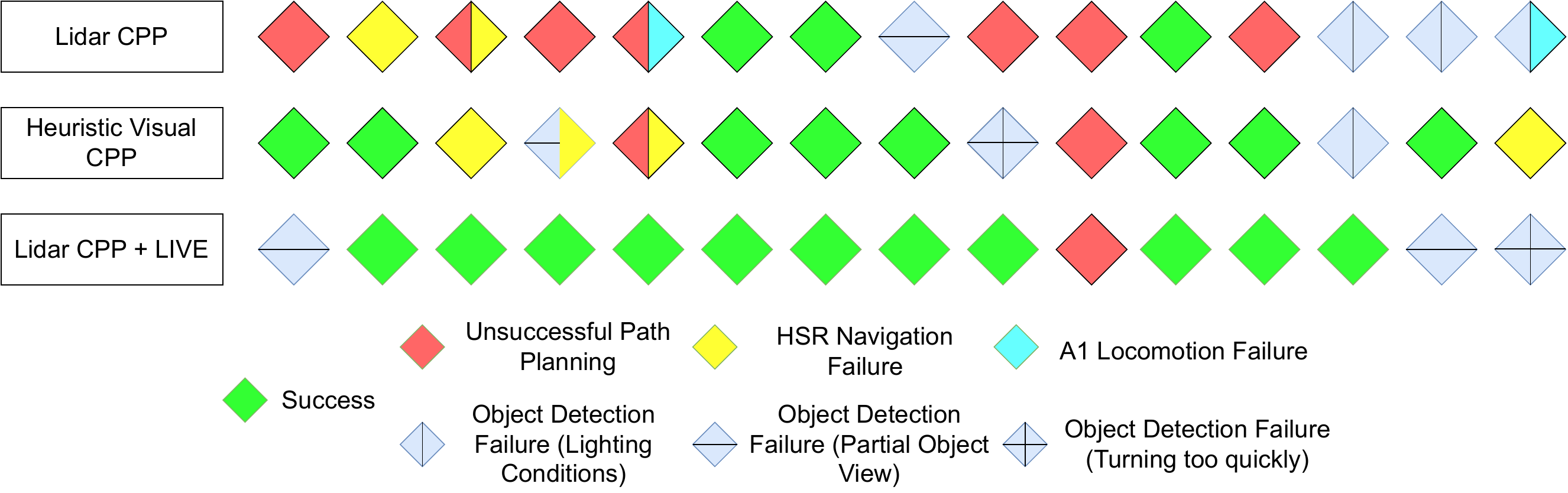}
    \captionsetup{belowskip=1pt}
    \caption{Success / Failure Modes for each of 15 trials in all three path planning variants. The addition of LIVE significantly improves overall success rate and reduces the frequency of HSR Navigation Failure with more efficient path lengths. }
    \label{fig:success}
\end{figure*}
Complete tabular results can be found on the project website, including trial-by-trial results, path lengths compared by initial condition, failure mode analysis, and success rate versus object difficulty.
Videos and robot trajectories for all trials can be found via the Google Drive link on the website. 
Figure \ref{fig:success} shows the success rate for each of the three planner settings as well as the relative frequency of each failure mode. 
A trial is classified `Path Failure' if the agents successfully follow the path plans output by the planner and failed to visually detect at least one object. 
`HSR Navigation Failure' indicates the HSR bumped into an object and triggered the emergency stop.
`A1 Locomotion Failure' occurs if the A1 falls while walking. 
`Object Detection Failure' occurs when at least a portion of the target object is in the RGB camera feed, yet the object detection algorithm misses it.
Notably, the occurrence of path failure in the Heuristic Visual CPP is similar to the proposed method, while Lidar CPP fails at a significantly higher rate, generating insufficient paths 40\% of the time.
Path lengths are computed in each setting with results from the proposed method and visual CPP in Table \ref{tab:lengths}, with Lidar CPP results on the website for brevity. 
In trials where one or more object is missed, the total lengths of the paths traversed by the robots is considered.
While Visual CPP rarely fails due to inadequate global paths, it suffers from `HSR Navigation Failures' in 16.7\% of trials. 
This can be attributed to the near 50\%, increase in path length over the other methods, resulting in higher localization drift and ultimately navigation failure.

\begin{figure}
    \centering
    \includegraphics[width=\linewidth]{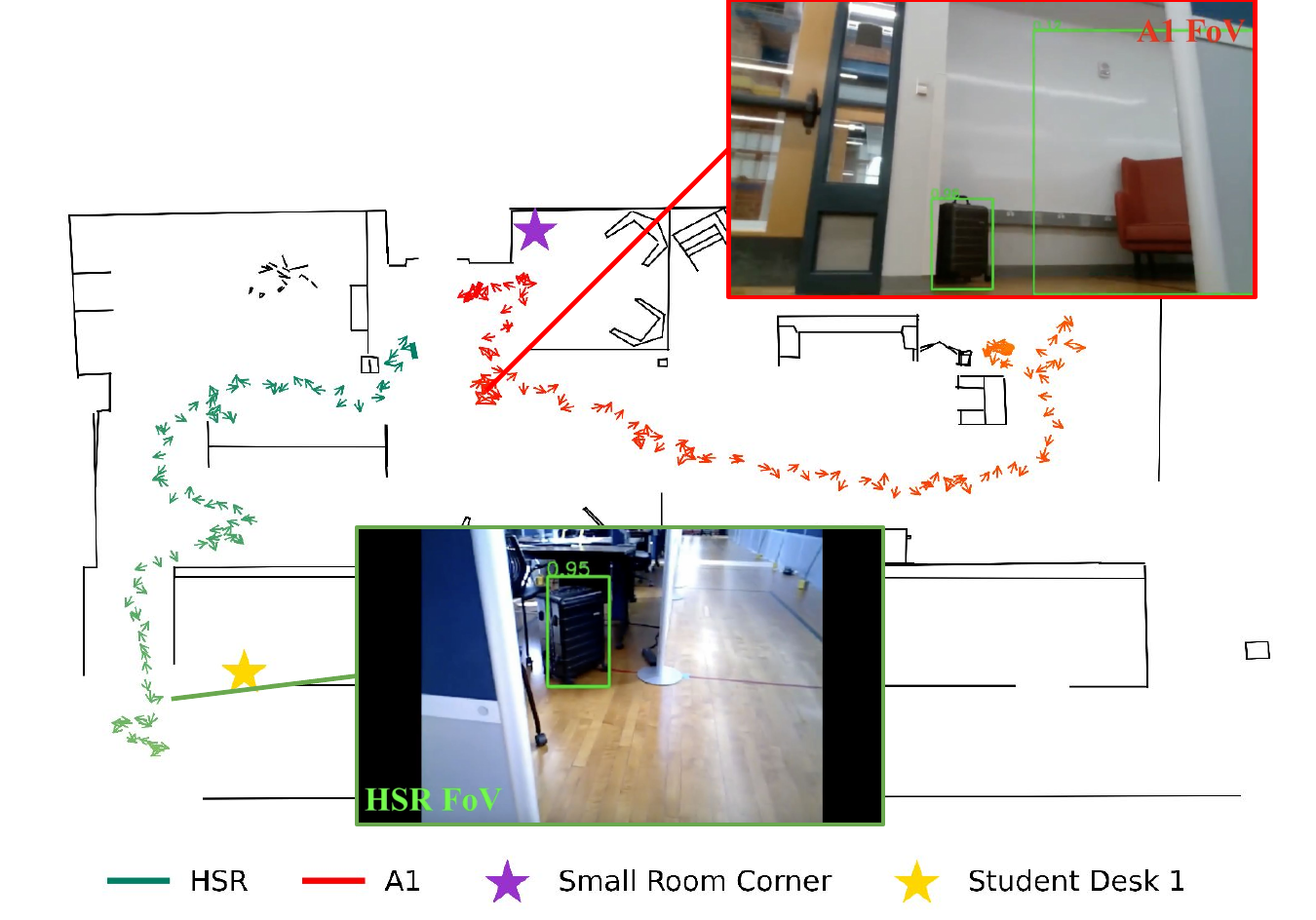}
    \caption{Top view of the static map with robot trajectories from LIVE trial 9 overlaid. Arrows represent robot pose as recorded using rosbag during execution. Included is the moment of object detection for each of the robots connected with the robot pose at that moment.\loosepar{} }
    \label{fig:ctrial9-trajectory}
\end{figure}
While the Heuristic Visual CPP path plans are successful, Lidar CPP + LIVE shows to be more robust in finding the objects in real-world scenarios due to this reduction in path length.
Trials throughout all three planning methods suffer from object detection shortcomings. 
In some cases, the robot turns too quickly for the algorithm to detect the suitcase. 
In others, detection was impacted by reflections of the sun and partial object views, however, the occurrence is nearly equal over the three planner settings. 
The results indicate the addition of LIVE improves real-world task performance with a success rate boost of 20\% as compared with Heuristic Visual CPP and 50\% as compared with the Lidar CPP baselines. \loosepar{}

%

We further inspect success rate for each method based on object difficulty. 
In each of the three settings, nine objects are located in Easy, 12 in Medium, and nine in Hard locations.
Fig. \ref{fig:map-ic-targets} depicts named object locations with color coded difficulty level.
Results in tabular form are omitted for brevity but can be found on the project website.
Easy object locations are found nearly 100\% of the time across all three experiment settings. 
When object locations are Medium or Hard, however, Lidar CPP baseline performs poorly, with combined success rate of 14.3\%.
This result is expected given that Lidar CPP does not account for the visual sensor.
In particular, Lidar CPP + LIVE detects 100\% of Medium objects and 67\% of Hard objects compared to Heuristic Visual CPP's 67\% and 44\%, respectively.
Priority waypoints from LIVE are shown to result in successful object detection in trials 4, 5, 8, 9, and 15.
Specifically, these trials have objects in Medium and Hard locations. 
Fig. \ref{fig:ctrial9-trajectory} is a representative trajectory plot from LIVE trial 9. 
Both objects in trial 9 are found with a priority waypoint and the moment of detection is overlaid on trajectories. 

Time data for the experiments can be found on the project website. 
The data displays a trend that LIVE improves path length efficiency and success rate, but must trade off detection time to explore inspection regions to achieve such success. 
This can be seen in particular in LIVE trial 13 before the HSR finds the object located `Behind Fridge.'
We note the following on comparing times for the three methods tested: 
1) Due to the relative success rates, the data is skewed towards those more difficult trials where the baseline methods are less successful, 
2) The variance of time data is high, indicating the initial conditions and object locations are well dispersed, and 
3) During some trials there are pauses in robot navigation, confounding the relationship between actual search time and quality of paths generated.\loosepar{}

\section{Conclusion} 
\label{sec:conclusion}
We present an algorithm that leverages efficient global path plans, 2D range data and map information to efficiently find objects in known environments.
We present results supporting that LIVE is more robust and efficient than global planning methods alone for real-world multi-object search. 
Ongoing work involves extending this method to unknown environments with supervised learning.
Further, incorporation of inspection regions in a utility function for viewpoint selection.\loosepar{}

\section*{Acknowledgments}
This research was supported in part by NSF Award \#2219236 and Living and Working with Robots, a core research project of Good Systems, a UT Grand Challenge.  Any opinions, findings, and conclusions or recommendations expressed in this material are those of the author and do not necessarily reflect the views of the National Science Foundation.
Research was in part sponsored by the Army Research Office and was accomplished under Cooperative Agreement Number W911NF-19-2-0333. 
 The views and conclusions contained in this document are those of the authors and should not be interpreted as representing the official policies, either expressed or implied, of the Army Research Office 
 or the U.S. Government. 
 The U.S. Government is authorized to reproduce and distribute reprints for Government purposes notwithstanding any copyright notation herein.


\bibliographystyle{plainnat}
\bibliography{references}

\end{document}